\documentclass[letterpaper, 10 pt, conference]{ieeeconf}  % Comment this line out if you need a4paper

\IEEEoverridecommandlockouts                              % This command is only needed if 
\usepackage{graphicx} % for pdf, bitmapped graphics files
\usepackage{amsmath} % assumes amsmath package installed
\usepackage{amsfonts,amssymb}
\usepackage[dvipsnames]{xcolor}
\usepackage{algorithm, algorithmic}
\usepackage{tabularx, booktabs, makecell, caption}
\usepackage{tabularray}
\usepackage{siunitx}
\usepackage{pifont}
\usepackage{url}
\usepackage{hyperref}

\makeatletter
\def\algbackskip{\hskip-\ALG@thistlm}
\makeatother

\title{\LARGE \bf
Grounding Language Models in Autonomous Loco-manipulation Tasks
}

\author{Jin Wang$^{1}$$^{*}$, Nikos Tsagarakis$^{1}$% <-this % stops a space
\thanks{†This work was supported by the European Union’s Horizon 2020 research and innovation programme, euROBIN EPUE034001 and Leonardo Joint Lab ETCM058502.}% <-this % stops a space
\thanks{$^{1}$Humanoids and Human-Centered Mechatronics (HHCM), Istituto Italiano di Tecnologia, Via Morego 30, Genoa, 16163, Italy.}
\thanks{$^{*}$Email: {\tt\small wang.jin@iit.it}}
}

\begin{document}

\maketitle
\thispagestyle{empty}
\pagestyle{empty}

%%%%%%%%%%%%%%%%%%%%%%%%%%%%%%%%%%%%%%%%%%%%%%%%%%%%%%%%%%%%%%%%%%%%%%%%%%%%%%%%
\begin{abstract}
Humanoid robots with behavioral autonomy have consistently been regarded as ideal collaborators in our daily lives and promising representations of embodied intelligence. Compared to fixed-based robotic arms, humanoid robots offer a larger operational space while significantly increasing the difficulty of control and planning. Despite the rapid progress towards general-purpose humanoid robots, most studies remain focused on locomotion ability with few investigations into whole-body coordination and tasks planning, thus limiting the potential to demonstrate long-horizon tasks involving both mobility and manipulation under open-ended verbal instructions. In this work, we propose a novel framework that learns, selects, and plans behaviors based on tasks in different scenarios. We combine reinforcement learning (RL) with whole-body optimization to generate robot motions and store them into a motion library. We further leverage the planning and reasoning features of the large language model (LLM), constructing a hierarchical task graph that comprises a series of motion primitives to bridge lower-level execution with higher-level planning. Experiments in simulation and real-world using the CENTAURO robot show that the language model based planner can efficiently adapt to new loco-manipulation tasks, demonstrating high autonomy from free-text commands in unstructured scenes.

\end{abstract}

\vspace{-0.5em}
%%%%%%%%%%%%%%%%%%%%%%%%%%%%%%%%%%%%%%%%%%%%%%%%%%%%%%%%%%%%%%%%%%%%%%%%%%%
%%%%%%%%%%%%%%%%%%%%%%%%%%%%Introduction%%%%%%%%%%%%%%%%%%%%%%%%%%%%%%%%%%%
%%%%%%%%%%%%%%%%%%%%%%%%%%%%%%%%%%%%%%%%%%%%%%%%%%%%%%%%%%%%%%%%%%%%%%%%%%%
\section{INTRODUCTION}

Maintaining autonomy during the execution of a task in a real-world environment is both essential and challenging for robots, especially when performing tasks that require interaction with surroundings and manipulation of objects. This involves robots being able to reason the given semantic instructions and plan their behaviors while using multi-modality to perceive and infer affordances and spatial geometric constraints of the environment to determine proper motions. 

Recently, the rise of large language models (LLMs) and their remarkable capabilities in robotic planning have made it possible to perform logical reasoning and construct hierarchical action sequences for complex tasks \cite{ichter2022do}. However, extending language model based algorithms to humanoid robots remains a challenge stemming from their complex dynamics and precise coordination between different components. 

In this study, we present a language model based framework enabling task reasoning and autonomous behavior planning towards humanoid loco-manipulation. 
We use a decomposed training strategy that modularly selects the components needed for specific tasks and maps low-dimensional space trajectories to the whole-body space with a unified motion generator. The trained actions are stored as skill primitive in a motion library. We adopt the LLM to decompose complex instructions consisting of multiple sub-tasks that select skills from the motion library and arranges a sequence of actions, referred to as a task graph. By leveraging the interaction of distilled spatial geometry and 2D observation with a visual language model (VLM) to ground knowledge into a motion morphology selector to choose appropriate actions in single- or dual-arm, legged or wheeled locomotion. We further illustrate through experiments how our framework can be learned and deployed on a high-DoF, hybrid wheeled-leg robot, performing zero-shot online planning under human instruction. Corresponding video can be seen here \cite{hymotion}.

\begin{figure}
\centerline{\includegraphics[width=8cm]{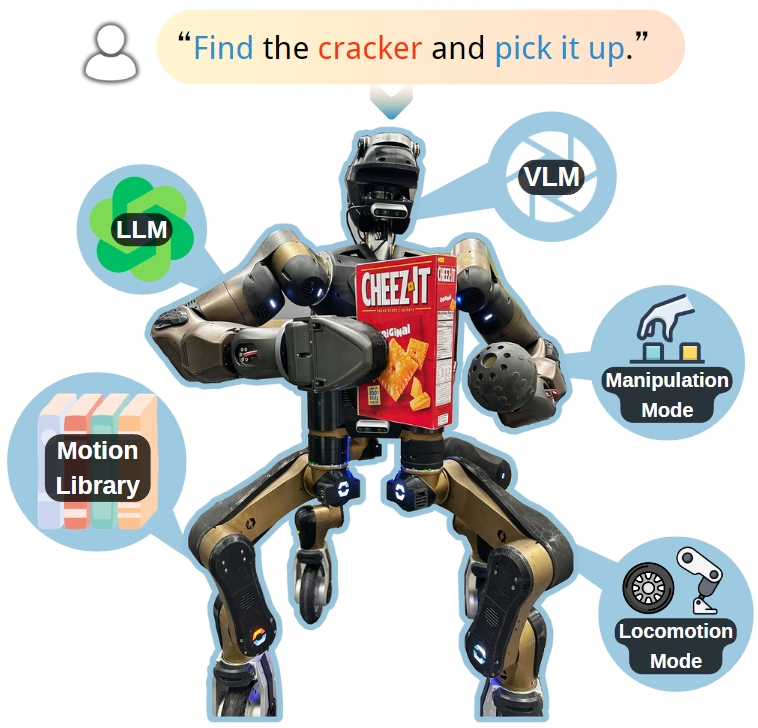}}
\caption{Humanoid robot CENTAURO picks objects with the planning of the $\textit{task graph}$ generated by the LLM. The `$\textit{Motion Library}$' consists of various learned behavior primitives. The VLM selects `$\textit{Manipulation Mode}$' and `$\textit{Locomotion Mode}$' based on different task scenarios.}
\vspace{-1.5em}
\label{fig1}
\end{figure}

 %, and ground the  LLM in humanoid robot to realize embodied intelligence.

%%%%%%%%%%%%%%%%%%%%%%%%%%%%%%%%%%%%%%%%%%%%%%%%%%%%%%%%%%%%%%%%%%%%%%%%%%%
%%%%%%%%%%%%%%%%%%%%%%%%%%%%Autonomous Behavior Planning%%%%%%%%%%%%%%%%%%%
%%%%%%%%%%%%%%%%%%%%%%%%%%%%%%%%%%%%%%%%%%%%%%%%%%%%%%%%%%%%%%%%%%%%%%%%%%%
\section{Methodology}

\begin{figure*}
 \centerline{\includegraphics[width=16cm]{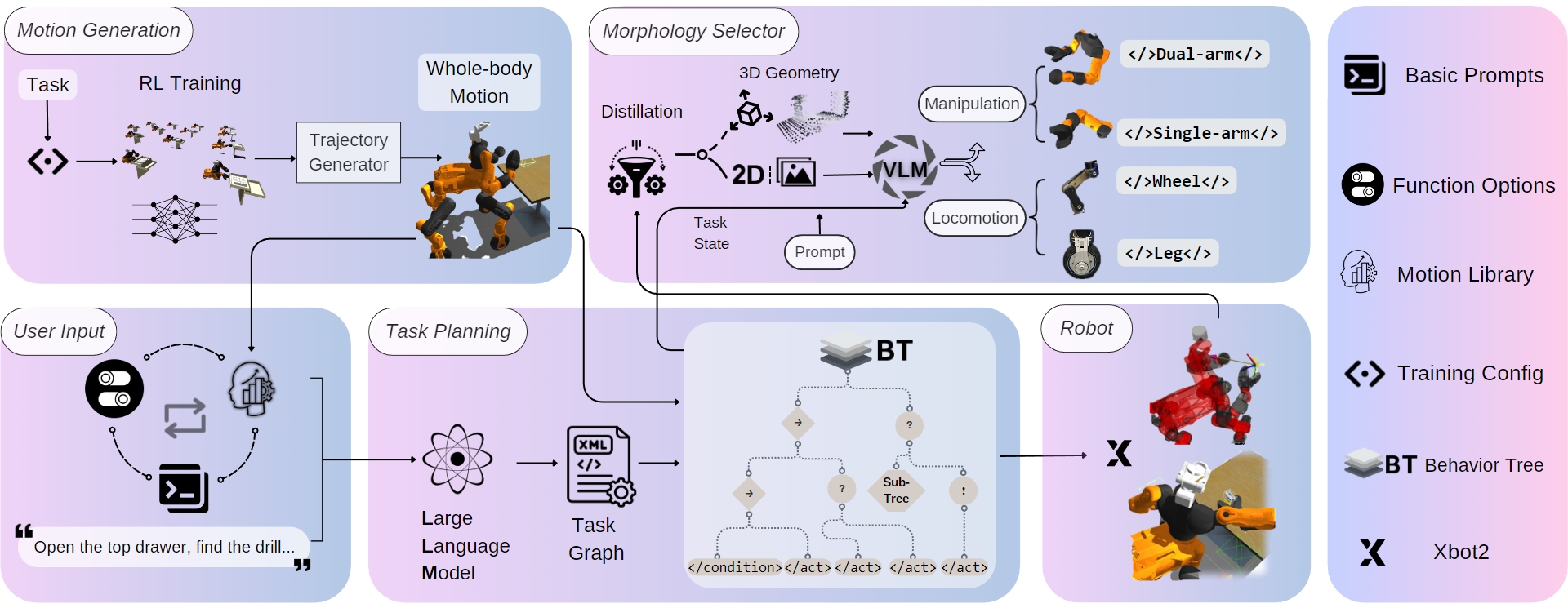}}
\caption{Overview of the Framework. \textbf{Motion generation} is assigned for learning and training whole-body motion skills for new tasks and storing them in the motion library. \textbf{User input} includes received task instructions and initialization prompt sets. \textbf{Task planning} generates a task graph that guides the robot's behavior and passes action commands to the real robot. \textbf{Morphology Selector} is used for motion mode determination in specific sub-tasks, selecting the appropriate morphology for locomotion and manipulation based on spatial affordances and robot intrinsic features.}
\vspace{-1.5em}
\label{framework}   
\end{figure*}

We illustrate how the proposed framework enables the humanoid robot CENTAURO \cite{centauro} to autonomously perform loco-manipulation guided by semantic instructions. As shown in Fig. 2, we divide the pipeline into four main interrelated sectors that are learned and deployed sim-to-real manner. The motion generation sector selects RL training configurations for specific tasks and conducts training in parallel. The trajectory obtained from the training is provided as a reference to the optimizer, which ultimately generates whole-body motion skills and the skills will be stored in the motion library. The user input sector contains a user interface as well as pre-defined basic prompts, function options, and motion library, all of which together constitute the textual material fed to the LLM. After receiving a command, the task planning sector generates a hierarchical task graph using the LLM. Once the task graph is loaded, it is interpreted as a Behavior Tree to guide the robot's execution. When a task requires selecting the motion morphology, depth-sensing information is invoked and distilled into 2D images and geometric features. These data, along with the task state and prompts, are fed to the VLM, which then selects the morphology capable of achieving the goal. Through the coordination of these sectors, the study facilitates semantic command understanding and zero-shot behavioral planning and action execution for CENTAURO robot.

%%%%%%%%%%%%%%%%%%%%%%%%%%%%%%%%%%%%%%%%%%%%%%%%%%%%%%%%%%%%%%%%%%%%%%%%%%%
%%%%%%%%%%%%%%%%%%%%%%%%%%%%Experiment and Evaluation%%%%%%%%%%%%%%%%%%%
%%%%%%%%%%%%%%%%%%%%%%%%%%%%%%%%%%%%%%%%%%%%%%%%%%%%%%%%%%%%%%%%%%%%%%%%%%%
\section{Experiment}

We validate the ability of LLM to plan motion primitives for different loco-manipulation tasks. Experiments were conducted on tasks requiring a combination of perceptions and actions. We recorded the success rate and the impact of different errors of 4 representative tasks and provided quantitative evaluations in Fig.8. The results show the LLM based planner can effectively plan for semantic instructions based on learned skills and guide the robot to complete a variety of tasks according to the action sequences, achieving a desired success rate ($\ge60\%$) on real-world robot. And adding failure detection and recovery (FR) to the planning increases the success rate of task execution. Whereas selecting multiple functional modules as input also increases the difficulty of planning, and execution errors mainly stem from intricate dynamical constraints on the actions and misalignment of the floating sensing with robot execution.

\section{Conclusion}
In this work, we present a novel framework that tackles behavior planning towards different tasks and enables humanoid robots to perform autonomously loco-manipulation tasks through grounded language model. By leveraging the interaction of distilled spatial geometry and 2D observation with VLM, it helps to choose motion morphology of the CENTAURO robot for various scenarios, and bridge the gap between semantic space, robotic perception and action. For further details of this work, please refer to \cite{wang2024hypermotion}.

\begin{figure}
\centerline{\includegraphics[width=9cm]{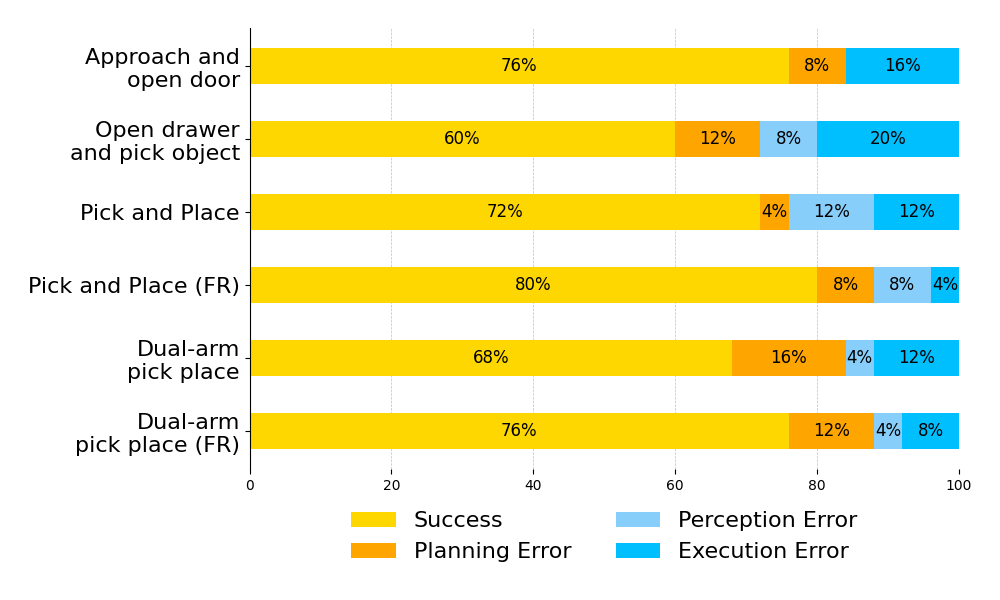}}
\caption{Average rate of CENTAURO robot successfully performing various LLM planning tasks, and failure caused by different type of errors during the tasks.}
\vspace{-1.5em}
\label{fig1}
\end{figure}

\bibliographystyle{IEEEtran}
\bibliography{references}

\addtolength{\textheight}{-12cm}   % This command serves to balance the column lengths

\end{document}